\documentclass[10pt,twocolumn,letterpaper]{article}
\usepackage[pagenumbers]{jopano}

\definecolor{jopanoblue}{rgb}{0.21,0.49,0.74}
\usepackage[pagebackref,breaklinks,colorlinks,allcolors=jopanoblue]{hyperref}
\usepackage{multirow}

\usepackage{utils}
\usepackage{tabularx}
\usepackage[misc,geometry]{ifsym}
\NewDocumentCommand{\chzl}{o m}{\comm[HZL Comment][red][#1]{#2}}
\NewDocumentCommand{\cfwc}{o m}{\comm[FWC Comment][magenta][#1]{#2}}
\NewDocumentCommand{\addfwc}{m o}{\add[violet][FWC Comment]{#1}[#2]}
\NewDocumentCommand{\strkfwc}{m o}{\strk[violet][FWC Comment]{#1}[#2]}
\NewDocumentCommand{\delfwc}{m o}{\del[violet][FWC Comment]{#1}[#2]}
\NewDocumentCommand{\rplfwc}{m m o}{\rpl[violet][FWC Comment]{#1}{#2}[#3]}

\newcommand{\MODEL}{JoPano\xspace}

\title{\MODEL: Unified Panorama Generation via Joint Modeling\vspace{-0.5ex}}

\author{
  Wancheng Feng$^{1,3\,\text{*}}$~~
  Chen An$^{1,2\,\text{*}}$~~
  Zhenliang He$^{1\>\text{\Letter}}$~~
  Meina Kan$^{1,2}$~~
  Shiguang Shan$^{1,2}$~~
  Lukun Wang$^{3}$\vspace{1.5ex}\\
  $^{1}$State Key Laboratory of AI Safety, Institute of Computing Technology, CAS, China\\
  $^{2}$University of Chinese Academy of Sciences (CAS), China\\
  $^{3}$Shandong University of Science and Technology, China\vspace{1.5ex}\\
  \url{https://VIPL-GENUN.github.io/Project-JoPano}
}

\begin{document}

\twocolumn[{
\maketitle
\vspace{-12mm}
\begin{center}
\includegraphics[width=\textwidth]{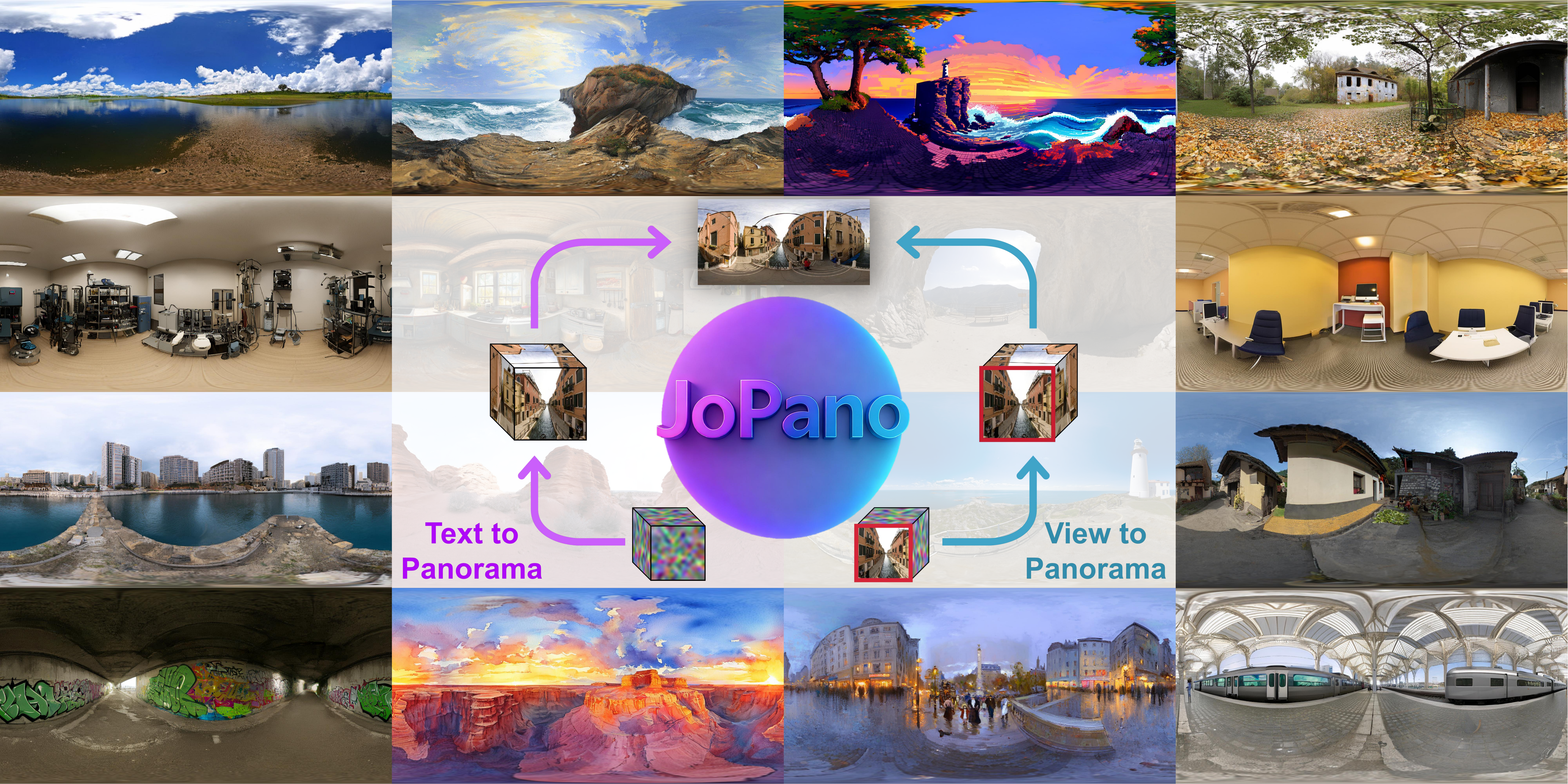}
\vspace{-6.5mm}
\captionof{figure}{We propose JoPano, a unified panorama generation framework that supports both text-to-panorama (T2P) and view-to-panorama (V2P). The left eight examples show T2P results, while the right eight show V2P results. JoPano generates high-quality panoramas across indoor, outdoor, and stylized scenes.}
\vspace{2.5mm}
\label{fig:teaser}
\end{center}
}]

\begingroup
\renewcommand\thefootnote{\fnsymbol{footnote}}%
\footnotetext{* Equal contribution. \Letter~Corresponding author.}%
\endgroup

\begin{abstract}
Panorama generation has recently attracted growing interest in the research community, with two core tasks, text-to-panorama and view-to-panorama generation.
However, existing methods still face two major challenges: their U-Net-based architectures constrain the visual quality of the generated panoramas, and they usually treat the two core tasks independently, which leads to modeling redundancy and inefficiency.
To overcome these challenges, we propose a \textbf{jo}int-face \textbf{pano}rama (\textbf{\MODEL}) generation approach that unifies the two core tasks within a DiT-based model.
To transfer the rich generative capabilities of existing DiT backbones learned from natural images to the panorama domain, we propose a Joint-Face Adapter built on the cubemap representation of panoramas, which enables a pretrained DiT to jointly model and generate different views of a panorama.
We further apply Poisson Blending to reduce seam inconsistencies that often appear at the boundaries between cube faces.
Correspondingly, we introduce Seam-SSIM and Seam-Sobel metrics to quantitatively evaluate the seam consistency.
Moreover, we propose a condition switching mechanism that unifies text-to-panorama and view-to-panorama tasks within a single model.
Comprehensive experiments show that \MODEL can generate high-quality panoramas for both text-to-panorama and view-to-panorama generation tasks, achieving state-of-the-art performance on FID, CLIP-FID, IS, and CLIP-Score metrics.
\end{abstract}    
\vspace{-8mm}
\section{Introduction}
\label{sec:intro}
\vspace{-1mm}
A panorama is a 2D representation of a scene that can cover the entire \(360^\circ\) field of view.
It has been widely adopted in interactive and immersive applications such as Virtual Reality and Augmented Reality~\cite{bai2025360,zhou2024dreamscene360}, and it has also emerged as a promising representation for World Models~\cite{team2025hunyuanworld,yu2024wonderjourney,yu2025wonderworld}.
However, the acquisition of real-world panorama data relies on specialized equipment~\cite{xiao2012recognizing,chang2017matterport3d,gardner2017learning}, making large-scale collection expensive and challenging.
Therefore, the synthesis of panoramas has emerged as a significant research focus~\cite{ye2024diffpano,zheng2025panorama,wupanodiffusion,zhang2024taming,huang2025dreamcube,kalischek2025cubediff}, particularly facilitated by recent advances in diffusion models~\cite{ho2020denoising,DBLP:conf/iclr/SongME21,DBLP:conf/iclr/LipmanCBNL23,rombach2022high,podell2023sdxl,peebles2023scalable,esser2024scaling,flux2024} for generating high-quality visual content.
Nevertheless, current panorama generation methods still encounter challenges related to generation quality and modeling efficiency.

\textit{Challenge 1: The visual quality of generated panoramas remains limited in terms of resolution and detail.}
Most existing methods are based on the U-Net architecture~\cite{rombach2022high,podell2023sdxl}, which limits their ability to generate high-quality results. 
Recently, diffusion transformers (DiT)~\cite{peebles2023scalable,flux2024,esser2024scaling,xie2024sana} have demonstrated strong generative capacity in the natural image domain, providing promising foundation models for panorama generation~\cite{ye2024diffpano,zheng2025panorama,wupanodiffusion,zhang2024taming,huang2025dreamcube,kalischek2025cubediff}.
However, most DiT backbones are designed for and pretrained on natural images, which exhibit a significant domain gap compared to panoramic images~\cite{huang2025dreamcube, kalischek2025cubediff}.
Therefore, a key problem arises: how to adapt a pretrained DiT from the natural image domain to the panorama image domain while maintaining its capability for high-quality generation as well as other practical functionalities such as stylized generation.

\textit{Challenge 2: Although the two core tasks of panorama generation are closely related, they have been developed independently, resulting in modeling redundancies and inefficiencies.}
Panorama generation comprises two core tasks: 1) text-to-panorama (T2P), where the model generates a panorama according to a textual description~\cite{ni2025makes, zhang2024taming, ye2024diffpano, yang2024dreamspace, wupanodiffusion, li2023panogen, tang2023MVDiffusion}, and 2) view-to-panorama (V2P), where the model completes a panorama given a narrow field of view~\cite{kalischek2025cubediff, huang2025dreamcube, zheng2025panorama, nakata20242s, wang2023360, akimoto2022diverse}.
Previous methods typically design specific solutions for each task independently.
However, since both tasks can be regarded as conditional generation problems within the panorama domain, they may share intrinsic commonalities in their generative mechanisms.
Therefore, we believe there is potential to integrate these two tasks under a unified paradigm, thereby reducing the number of models and improving modeling efficiency.

In this paper, we introduce \textbf{\MODEL} to address the aforementioned challenges, achieving high-quality and efficient panorama generation.

\textit{For Challenge 1:} To improve the generation quality, we choose Sana~\cite{xie2024sana}, an efficient DiT architecture, as our backbone model.
We represent a panorama as six cube faces using a cubemap projection~\cite{greene1986environment,kalischek2025cubediff,huang2025dreamcube}, where each face corresponds to a perspective image of the scene.
Then, our goal is to learn to simultaneously generate all cube faces.
The core problem is: how to model the relationship and coherence among the six faces based on the pretrained Sana backbone.
To this end, we introduce \textit{Joint-Face Adapter} modules into the backbone, which apply normalization and full attention across all cube faces to jointly model their features and facilitate their interaction. 
Besides, we optimize only the adapter modules without altering the parameters of the Sana backbone, thereby preserving its original capabilities, such as high-quality generation and stylized generation.
Furthermore, to mitigate the seam inconsistencies that often appear at the boundaries between cube faces~\cite{huang2025dreamcube}, we apply Poisson Blending~\cite{perez2003poisson}, which effectively smooths the transitions between adjacent faces.
Correspondingly, we introduce Seam-SSIM and Seam-Sobel metrics to quantitatively evaluate the seam consistencies.
Overall, we achieve high-quality and seamless panorama generation.

\textit{For Challenge 2:} To reduce modeling redundancy and enhance efficiency, we propose a condition switching mechanism that unifies T2P and V2P tasks within a single diffusion model.
This mechanism allows the model to flexibly switch between the two tasks by changing only the conditioning inputs.
Under this unified setting, for the T2P task, the model simultaneously generates all six cubemap faces based on a text condition, where all faces are initialized with noise and denoised through a diffusion process.
For the V2P task, one cubemap face is provided as a view condition, and the model generates the remaining five faces from noise under the same diffusion process. 
Correspondingly, the diffusion loss is computed on all six faces for T2P and on the five generated faces for V2P during training.
In this manner, JoPano can flexibly and efficiently switch between the two tasks, eliminating the redundancy of separate~modeling.

Our contributions can be summarized as follows:
\begin{itemize}
    \item We propose a Joint-Face Adapter to transfer the generative capabilities of Sana-DiT from the natural image domain to the panorama domain, achieving high-quality and style-rich panorama generation.
    
    \item We unify text-to-panorama and view-to-panorama generation within a single diffusion framework via a condition-switching mechanism, enhancing the modeling efficiency.
    
    \item Our model achieves the state-of-the-art performance in panorama generation, surpassing existing methods in both visual quality and quantitative evaluations.
\end{itemize}

\section{Related Work}
\label{sec:related}

\subsection{Panorama Representations}
Panoramas are typically represented using two projection formats: the \textit{equirectangular projection} (ERP) and the \textit{cubemap projection} (CMP).

\paragraph{Equirectangular Projection}
The ERP projects a spherical panorama onto a 2:1 rectangular image, where the horizontal and vertical axes correspond to longitude and latitude, respectively.
Due to its simplicity, compact storage, and compatibility with standard image formats, ERP has been widely adopted for representing \(360^\circ\) panoramas~\cite{chen2022text2light,bai2025360,li2023panogen,zhou2025dense360,zhou2024dreamscene360,tang2023MVDiffusion,zhang2024taming,sun2025spherical,ni2025makes,wang2025conditional,akimoto2022diverse,dastjerdi2022guided,wang2022stylelight,lu2024autoregressive,zheng2025panorama}.

\paragraph{Cubemap Projection}
The CMP projects the spherical panorama onto the six square faces of a cube, each showing a \(90^\circ\) field of view in a different direction.
To mitigate the domain gap introduced by ERP, recent studies have explored panorama generation using cubemap~\cite{song2023roomdreamer,  ye2024diffpano, kalischek2025cubediff, huang2025dreamcube}.

\subsection{Panorama Generation}
Current panorama generation methods can be categorized into text-to-panorama and view–to-panorama approaches.
Early text-to-panorama methods~\cite{chen2022text2light, may2023cubegan} relied on GANs~\cite{goodfellow2014generative}, while diffusion models~\cite{wang2024customizing, feng2023diffusion360} have recently enabled more sophisticated panorama generation.
PanoGen~\cite{li2023panogen} and MVDiffusion~\cite{tang2023MVDiffusion} generate panoramas via recursive outpainting based on pretrained text-to-image diffusion models.
DiffPano~\cite{ye2024diffpano} instead generates panoramas directly by fine-tuning a diffusion model with LoRA.
However, the outputs of these methods still suffer from geometric distortions due to the domain gap between natural images and panorama images.
To alleviate this issue, some methods~\cite{zhang2024taming, tan2024imagine360, park2025spherediff} adopt dual-branch architectures within diffusion models, which partially mitigate distortions but introduce substantial computational overhead. 
SMGD~\cite{sun2025spherical} introduces a spherical manifold convolution to guide the diffusion process and reduce distortion.
In contrast, PAR~\cite{wang2025conditional} employs an autoregressive model~\cite{dengautoregressive} to avoid the misalignment between ERP representations and diffusion models.
For the view-to-panorama task, early works typically follow an outpainting paradigm~\cite{akimoto2022diverse, dastjerdi2022guided, wang2022stylelight, lu2024autoregressive}.
More recent works~\cite{song2023roomdreamer, zhang2023diffcollage,  kalischek2025cubediff, huang2025dreamcube} leverage cubemap representations instead of ERP to mitigate geometric distortions. Among them, CubeDiff~\cite{kalischek2025cubediff} generates the six perspective views of a cubemap in parallel, while DreamCube~\cite{huang2025dreamcube} proposes a multi-plane synchronization strategy to alleviate discontinuous seams and inconsistent color tones in cubemap panorama generation.
Nevertheless, these methods still inevitably produce seam inconsistencies along cubemap face boundaries.
Despite these advances in projection design and task formulation, most panorama generation methods still rely on the U-Net-based architecture as the backbone, which limits the overall generation quality.
A concurrent work, HunyuanWorld~\cite{team2025hunyuanworld}, addresses this limitation by building on Flux~\cite{flux2024} and training with large-scale data, achieving higher-fidelity panorama~generation.

\subsection{Diffusion Model}

Diffusion models~\cite{ho2020denoising, DBLP:conf/nips/SongE19, DBLP:conf/iclr/SongME21, DBLP:conf/iclr/LipmanCBNL23,dhariwal2021diffusion,ramesh2022hierarchical,DBLP:conf/nips/KarrasAAL22,ho2022classifier} are powerful generative methods that synthesize data by reversing a process that gradually adds noise.
To further improve efficiency, Latent Diffusion Models~\cite{rombach2022high} perform the diffusion process in a compact latent space, significantly reducing computational cost while enabling high-quality generation. 
In terms of model architecture, the DiT-based~\cite{peebles2023scalable, flux2024, esser2024scaling, DBLP:conf/eccv/MaGABVX24,xie2024sana} has recently outperformed traditional U-Net-based designs~\cite{rombach2022high, podell2023sdxl, DBLP:conf/eccv/SauerLBR24, DBLP:conf/nips/SahariaCSLWDGLA22} in both scalability and generative quality.

\section{Method}
\label{sec:method}

\begin{figure*}[ht]
    \centering
    \includegraphics[width=\textwidth]{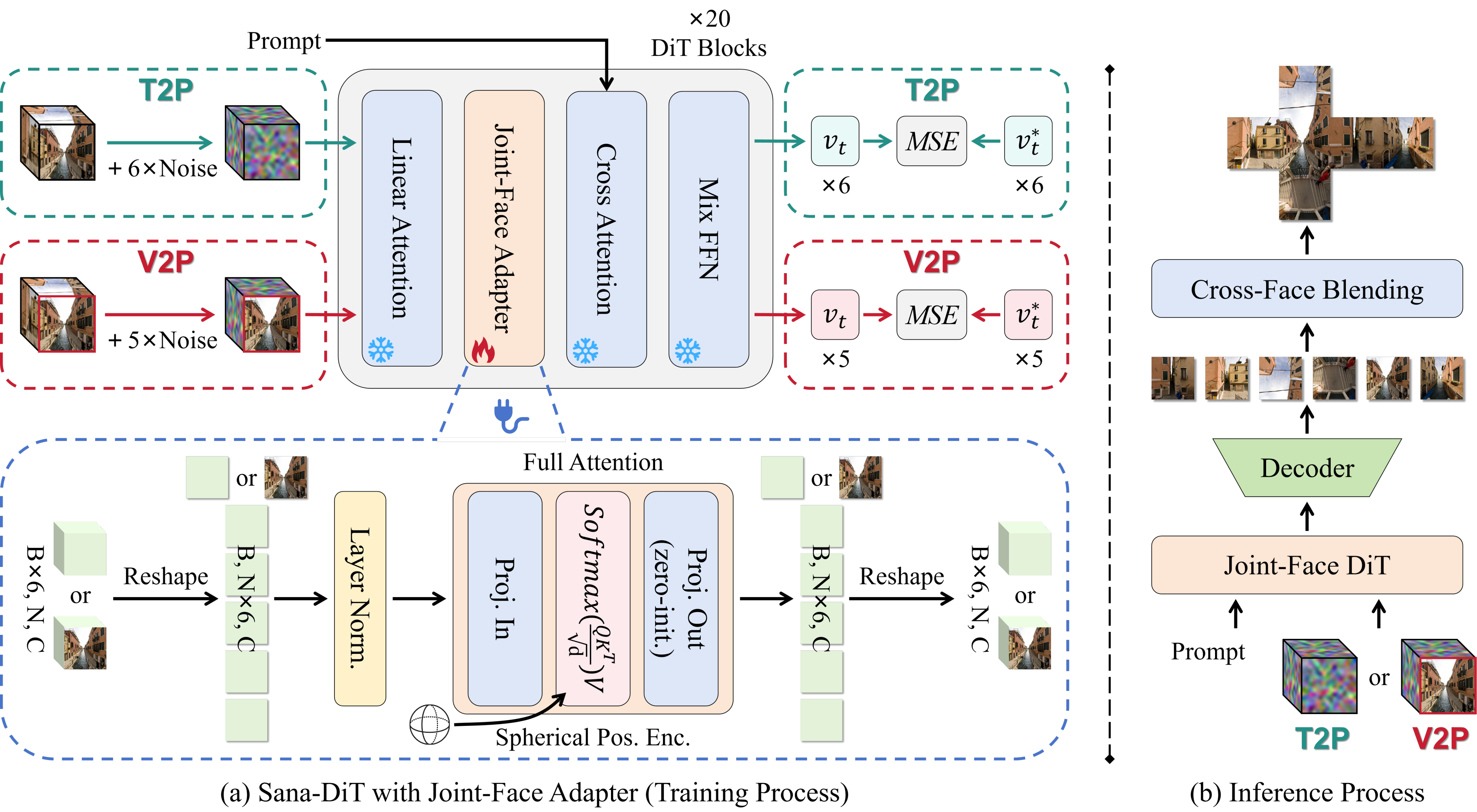}
    \caption{Overview of the JoPano pipeline. (a)Training process. The Joint-Face Adapter is inserted into Sana-DiT to jointly model all six cubemap faces, and a single diffusion process is shared by T2P and V2P. (b) Inference process. The Joint-Face DiT generates the cubemap faces, and the Cross-Face Blender further refines the results across faces.}
    \label{fig:main}
\end{figure*}

We represent a panorama as six cubemap faces $\{f_{i}\}_{i=0}^5$~\cite{greene1986environment,kalischek2025cubediff,huang2025dreamcube}. 
For T2P task, the model generates all faces from six noise $\{\epsilon_i\sim \mathcal{N}(0,1)\}_{i=0}^5$ and a text prompt $c_{text}$:
\begin{equation}
\{f_{i}\}_{i=0}^5 = G(\epsilon_0,\epsilon_1,\ldots,\epsilon_5,c_{text}).
\label{eq:t2p}
\end{equation}
For V2P task, one face $f_0$ is given as a view condition, and the remaining five faces are synthesized from $f_0$, five noise $\{\epsilon_i\sim \mathcal{N}(0,1)\}_{i=1}^5$, and $c_{text}$:
\begin{equation}
\{f_{i}\}_{i=0}^5 = G(\underline{f_0}, \epsilon_1, \ldots, \epsilon_5,c_{text}).
\label{eq:v2p}
\end{equation}
In the following sections, we introduce Joint-Face Adapter to extend a DiT model to the panorama domain and a condition switching mechanism to equip a single diffusion model with both T2P and V2P capabilities. The overall pipeline is shown in ~\cref{fig:main}.
%

\subsection{Joint-Face Adapter}
\label{sec:JFA}
To enforce cross-face interaction, we introduce the Joint-Face Adapter, which extends Sana~\cite{xie2024sana} to the panorama domain.
Specifically, we insert a Joint-Face Adapter into each DiT block, after the original linear attention and before the cross-attention layer, enabling the model to process all six cubemap faces simultaneously.

\paragraph{Joint Modeling}
We treat each cubemap face as an individual image in the DiT backbone.
Given tokens of all faces as $\mathbf{z}\in\mathbb{R}^{(B\times 6)\times N\times C}$,
where $B$ is the number of panoramas in a batch, $N$ is the number of tokens per face,
and $C$ is the channel dimension, we reshape them into $\hat{\mathbf{z}}\in\mathbb{R}^{B\times (N\times6)\times C}$ to concatenate the six cube faces of each panorama along the token dimension.
We first apply Layer Normalization~\cite{ba2016layer} to $\hat{\mathbf{z}}$, and denote the output as $\mathbf{y}\in\mathbb{R}^{B\times (N\times 6)\times C}$. The normalization is shared across all six faces.
To explicitly model cross-face interactions, we then apply full attention over the token sequence $\mathbf{y}$, allowing each token to attend to all tokens of all faces.
We obtain the Query, Key, and Value via linear projections of the normalized features $\mathbf{y}$:
\begin{equation}
Q = W_Q \mathbf{y},\quad
K = W_K \mathbf{y},\quad
V = W_V \mathbf{y}.
\end{equation}
where $W_Q$, $W_K$, and $W_V$ are projection matrices, and then the full attention is computed as
\begin{equation}
\hat{\mathbf{z}}'
= \operatorname{softmax}\!\left(\frac{QK^{\top}}{\sqrt{d}}\right)V.
\label{eq:fullattn}
\end{equation}
The output $\hat{\mathbf{z}}'\in\mathbb{R}^{B\times (N\times 6)\times C}$ is reshaped back to $\mathbf{z}'\in\mathbb{R}^{(B\times 6)\times N\times C}$, so that the updated features can be fed back into the pretrained diffusion backbone without changing its original shape.

\paragraph{3D Spherical Positional Embedding}
To better match the geometry of panoramas, we use the unit sphere for position embedding.
For each token on the cubemap faces, we place a unit sphere inside the cube and take the intersection of the line from the cube center to the token’s location on the face as its 3D coordinate $(x,y,z)$.
We then feed this unit 3D direction into RoPE~\cite{su2024roformer} to embed positional information into the queries and keys.

\paragraph{Adapter Only Optimization}
During training, we freeze the pretrained diffusion backbone and train only the Joint-Face Adapter, with its output projection zero-initialized.
This design makes the model behave like the pretrained diffusion backbone at the start of training, and then gradually adapt to panorama generation as the adapter learns cross-face dependencies.

\subsection{Unified Generation}
\label{sec:unifiedmodel}

\paragraph{Condition Switching}
During training, following Rectified Flow~\cite{DBLP:conf/iclr/LiuG023}, the model predicts the velocity field from cubemap faces perturbed with Gaussian noise and the given condition. For each face $f_i$, the noisy sample at timestep $t$ is defined as
\begin{equation}
f_{i,t} = (1-t)f_i+t\epsilon,\quad\epsilon \sim \mathcal{N}(0,1),\quad t\in(0,1).
\end{equation}

To train a single model for both T2P and V2P, we introduce a condition switching mechanism that selects the input configuration for each training sample. 
Specifically, we set a binary switch $\gamma\!\in\!\{0,1\}$, applied per training sample. 
In training, we set $\gamma=0$ (T2P) or $\gamma=1$ (V2P) with equal probability (0.5 / 0.5), ensuring that both tasks are balanced.
When $\gamma=0$, the model is conditioned only on the text $c_{text}$ and predicts the velocity field as
\begin{equation}
v_\theta(f_{0,t},f_{1,t},\ldots,f_{5,t},t, c_{text}).
\end{equation}
When $\gamma=1$, the model is conditioned on a fixed face $f_0$ and the text $c_{text}$ and predicts the velocity field as
\begin{equation}
v_\theta(\underline{f_0},f_{1,t},\ldots,f_{5,t},t, c_{text}).
\end{equation}
This fixed-face design avoids sampling over different faces and matches our implementation.

\paragraph{Loss Function}
We optimize the model with an MSE loss over the supervised faces:

\begin{equation}
\mathcal{L}
= \mathbb{E}_{t,\gamma}
\bigg[
\frac{1}{6-\gamma}
\sum_{i=\gamma}^{5}
\big\|
v_\theta^{(i)} - v^{*(i)}
\big\|_2^2
\bigg],
\end{equation}
where $v_\theta^{(i)}$ denotes the predicted velocity for face $f_i$, and $v^{*(i)} = \epsilon - f_i$ is the corresponding ground truth velocity. When $\gamma=0$, all six faces are supervised; when $\gamma=1$, $f_0$ serves as a clean view condition and remaining five faces are supervised.

\subsection{Cross-Face Seam Handling}
\label{sec:CFB}

\paragraph{Cross-Face Blending}
To alleviate seam inconsistencies between cubemap faces, we apply Poisson Blending~\cite{perez2003poisson}, which obtains the composite image by solving a Poisson equation with Dirichlet boundary conditions:
\begin{equation}
\Delta f = \text{div}\,\mathbf{v} \text{ in}\ \Omega,\ \text{with} f|_{\partial\Omega} = f^*|_{\partial\Omega}.
\end{equation}

In our setting, we apply it independently to each cubemap face $g_i$ .
For face $g_i$, let $\Omega_i$ denoteits image domain and $\mathbf{v}_i$ the gradient field.
For every edge shared by $g_i$ and its neighboring face $g_j$, we extract from both faces a narrow band of width one pixel along the edge, and set the Dirichlet boundary values on $\partial\Omega_i$ to the pixelwise average of these two bands.

\begin{equation}
\begin{cases}
\Delta f_i = \text{div}\,\mathbf{v}_i & \text{in } \Omega_i,\\[2pt]
f_i = \tfrac12\big(g_i + g_j\big) & \text{on } \partial\Omega_i,
\label{eq:poisson}
\end{cases}
\end{equation}
where $j=nbr(i,e)$, $e\in\{N,S,E,W\}$, denotes the face adjacent to $g_i$ in the direction $e$. 
The solution $f_i$ serves as the blended version of face $g_i$, yielding a cubemap with significantly reduced cross-face seams.

\paragraph{Seam Consistency Metrics}

To quantify seam consistencies between cubemap faces, we introduce two metrics: Seam-SSIM, based on SSIM~\cite{wang2004image}, and Seam-Sobel, based on image gradients.

For each cube edge $e \in \{1,\dots,12\}$, we take narrow boundary bands $B^{(L)}_e$ and $B^{(R)}_e$(width $1\%$ of the face) from the two faces on either side of the edge and Seam-SSIM is defined as
\begin{equation}
\text{Seam-SSIM}
= \frac{1}{12}\sum_{e=1}^{12}
\text{SSIM}\bigl(B^{(L)}_e, B^{(R)}_e\bigr),
\label{eq:seam_ssim}
\end{equation}
where higher values indicate better seam consistency.

For Seam-Sobel, we apply an x-direction Sobel operator to the two faces and take the column of Sobel gradients closest to the shared edge on each side, denoted by $c^{(L)}_e$ and $c^{(R)}_e$. We define 
\begin{equation}
\text{Seam-Sobel}
= \frac{1}{12} \sum_{e=1}^{12}
\frac{\operatorname{mean}\lvert c_e^{(L)}\rvert
      + \operatorname{mean}\lvert c_e^{(R)}\rvert}{2},
\label{eq:seam_sobel}
\end{equation}
where lower values indicate better seam consistency.

\begin{figure*}[ht]
    \centering
    \includegraphics[width=\textwidth]{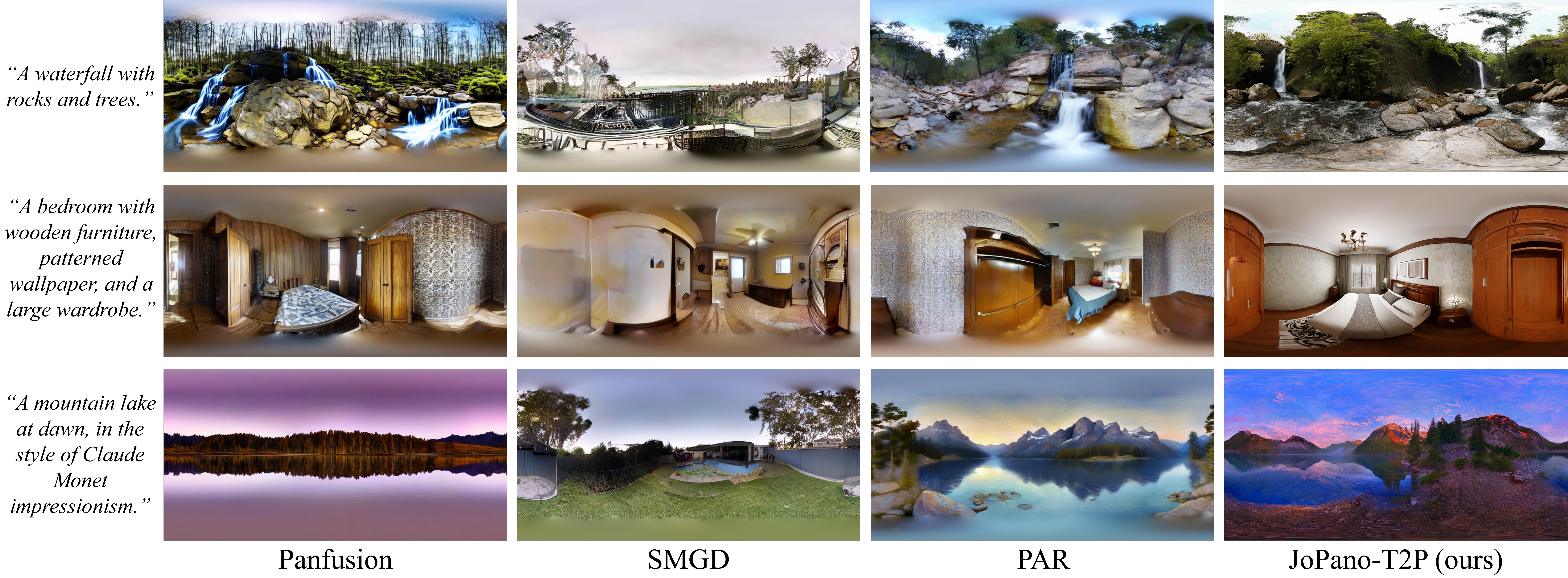}
    \caption{Comparison of JoPano with other T2P methods. 
    The first row shows outdoor scenes, the second row shows indoor scenes, and the third row shows a stylized scene, all generated from text prompts.}
    \label{fig:T2P compare}
\end{figure*}

\begin{figure*}[ht]
    \centering
    \includegraphics[width=\textwidth]{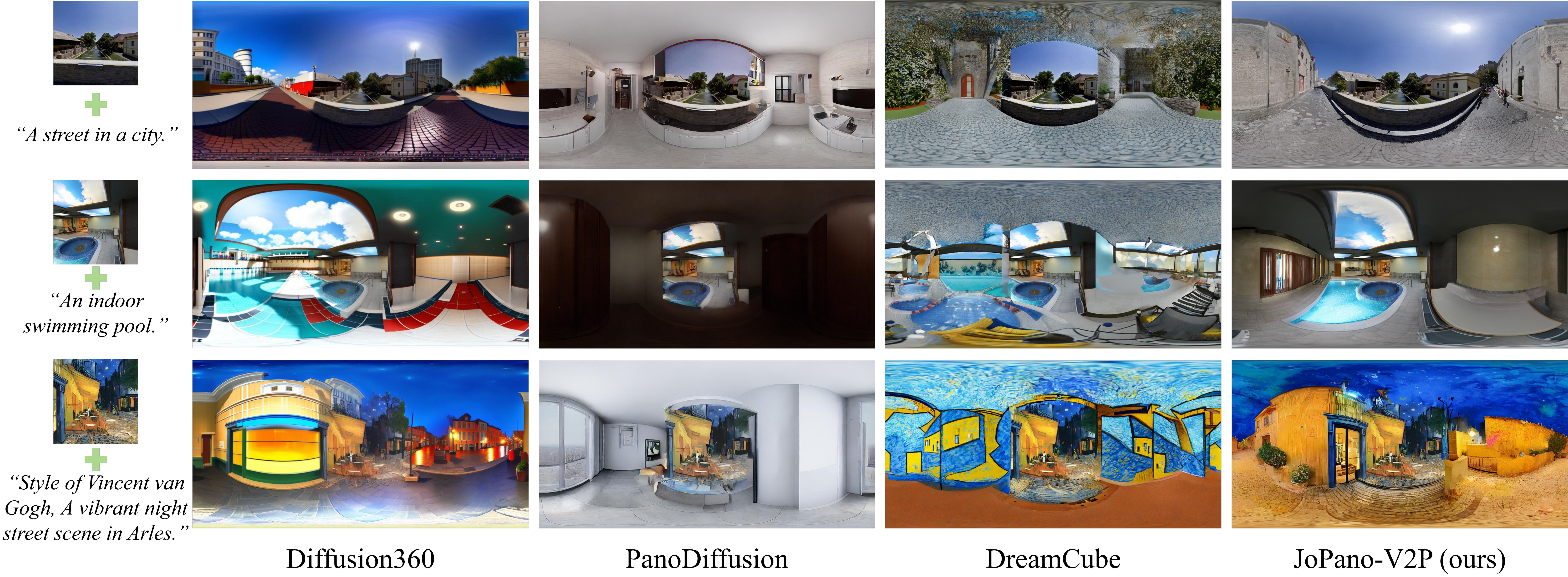}
    \caption{Comparison of JoPano with other V2P methods. 
    The first row shows outdoor scenes, the second row shows indoor scenes, and the third row shows a stylized scene, all generated from view conditions.}
    \label{fig:V2P compare}
\end{figure*}
\section{Experiment}
\label{sec:experimenr}

\subsection{Experimental Setup}

\paragraph{Implementation Details}
We use SANA-1.5 (1.6B) as the base model; after adding the Joint-Face Adapter, JoPano contains about 2B parameters. We train with a learning rate of $1\times10^{-4}$ for 1M steps on 8 Nvidia RTX A100-40G GPUs with a batch size of 1 per GPU. We adopt a cubemap resolution of $512\times512\times6$ and convert the generated cubemaps to ERP panoramas at $2048\times1024$ for visualization.

\paragraph{Dataset}
We use the Structure3D~\cite{zheng2020structured3d} and SUN360~\cite{xiao2012recognizing} datasets for training, containing 41{,}930 panoramas in total. 
Following~\cite{huang2025dreamcube,kalischek2025cubediff, wupanodiffusion}, we divide the Structure3D dataset into 16{,}930 panoramas for training, 2{,}116 for validation, and 2{,}117 for testing, and we use Qwen2.5-VL~\cite{bai2025qwen2} to caption each panorama.
For SUN360, we adopt the version provided by PanoDecouple~\cite{zheng2025panorama}, which contains 25{,}000 training and 4{,}260 testing panoramas paired with their corresponding text descriptions.
We use the test set of 2,117 panoramas from Structure3D (mostly indoor scenes) and 4,260 panoramas from SUN360 (mostly outdoor scenes) to evaluate the performance of T2P and V2P, respectively.

\begin{table*}[ht]
\centering
\caption{Quantitative comparison on SUN360 and Structure3D in terms of FID, CLIP-FID (CF), IS, and CLIP-Score (CS). For the T2P task, our method achieves state-of-the-art results on all metrics except the IS score on Structure3D, while for the V2P task it achieves state-of-the-art results on all metrics.}
\begingroup
\begin{tabularx}{\textwidth}{l l *{4}{>{\centering\arraybackslash}X} *{4}{>{\centering\arraybackslash}X}}
\toprule
\multirow{2}{*}[-0.7ex]{Task} & \multirow{2}{*}[-0.7ex]{Methods} & \multicolumn{4}{c}{SUN360} & \multicolumn{4}{c}{Structure3D} \\
\cmidrule(lr){3-6}\cmidrule(lr){7-10}
& & FID $\downarrow$ & \mbox{CF} $\downarrow$ & IS $\uparrow$ & \mbox{CS} $\uparrow$
  & FID $\downarrow$ & \mbox{CF} $\downarrow$ & IS $\uparrow$ & \mbox{CS} $\uparrow$ \\
\midrule
\multirow{4}{*}{T2P} 
 & PanFusion        & 30.92 & 23.76 & 7.12 & 29.62 & 48.53 & 23.34 & \textbf{3.97} & 22.31  \\
 & SMGD             & 48.87 & 38.85 & 4.77 & 17.09 & 53.34 & 27.97 & 3.37 & 23.73           \\
 & PAR              & 33.60 & 20.90 & 6.57 & 28.77 & 52.27 & 27.96 & 3.76 & 27.39           \\
 & JoPano (ours)    & \textbf{29.83} & \textbf{10.95} & \textbf{7.80} & \textbf{30.12} & \textbf{34.44} & \textbf{16.17} & 3.51 & \textbf{27.96}  \\
\midrule
\multirow{4}{*}{V2P} 
 & Diffusion360     & 33.12  & 24.88 & 6.34 & 25.85 & 36.68 & 13.83 & 2.67 &  25.28          \\
 & PanoDiffusion    & 125.33 & 37.42 & 3.48 & - & 22.47 & 7.78  & 3.00 & -           \\
 & DreamCube        & 43.83  & 17.02 & 4.80 & - & 25.10 & 6.89  & 2.84 & -          \\
 & JoPano (ours)    & \textbf{13.07} & \textbf{4.06} & \textbf{7.05} & \textbf{27.93} & \textbf{16.75} & \textbf{3.97} & \textbf{3.04} & \textbf{27.33}  \\
\bottomrule
\end{tabularx}

\endgroup
\label{tab:compare}
\end{table*}

\paragraph{Evaluation Metrics}
We evaluate our method using six metrics. To assess image quality, we report FID~\cite{heusel2017gans}, CLIP-FID (CF), and IS~\cite{salimans2016improved}. To evaluate text–image alignment, we use CLIP-Score (CS)~\cite{DBLP:conf/icml/RadfordKHRGASAM21}. In addition, we use Seam-SSIM and Seam-Sobel to evaluate seam consistency, but we only report these two metrics when comparing with DreamCube~\cite{huang2025dreamcube}, since they are specifically designed for cubemap panorama generation.

\begin{figure}[ht]
    \centering
    \includegraphics[width=\linewidth]{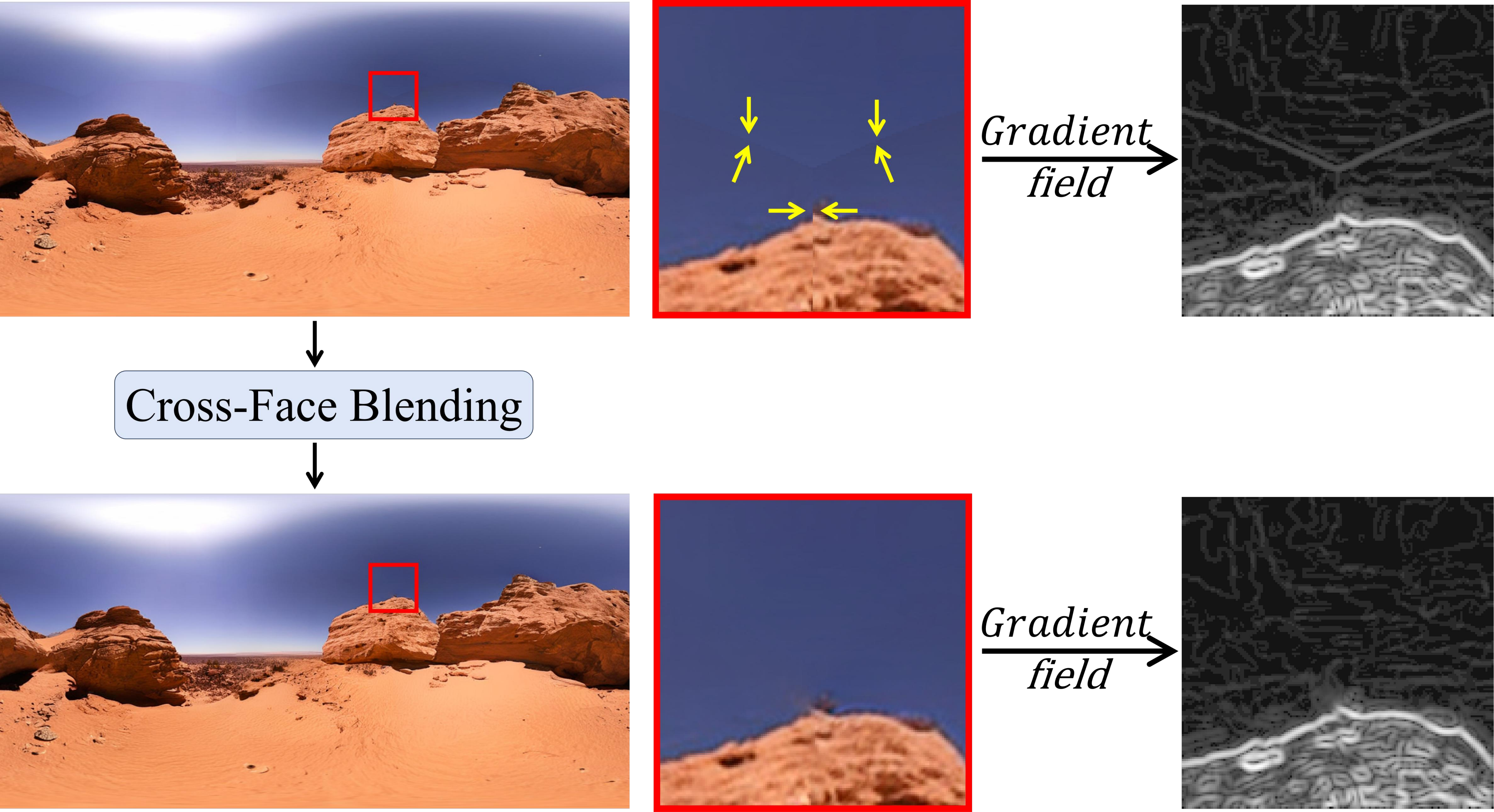}
    \caption{Comparison of ERP panoramas with and without Cross-Face Blending (CFB). The first row shows the panorama before CFB, with visible seam artifacts (red box). The second row shows the panorama after applying CFB, where the seams are smoothed. The improvement is more apparent in the gradient fields.}
    \label{fig:CFB}
\end{figure}

\subsection{Compare With Other Works}

\label{sec:compare_with_others}
We perform both quantitative and qualitative comparisons.
In the T2P setting, we compare JoPano with PanFusion~\cite{zhang2024taming}, SMGD~\cite{sun2025spherical}, and PAR~\cite{wang2025conditional}.
In the V2P setting, we compare with Diffusion360~\cite{wang2023360}, PanoDiffusion~\cite{wupanodiffusion}, and DreamCube~\cite{huang2025dreamcube}.
PanoDiffusion is an outpainting method, so its input consists only of a view image without any text prompt.
For DreamCube, we follow the configuration in the original paper and feed the model with a view image, its corresponding depth map, and six separate text prompts as inputs.

\paragraph{Quantitative Results}
We quantitatively compare our method with several panorama generation approaches.
We evaluate generation quality using FID, CLIP-FID, IS, and CLIP-Score.
As shown in \cref{tab:compare}, JoPano consistently outperforms existing methods on both SUN360 and Structure3D, achieving the best FID, CLIP-FID, and CLIP-Score for both T2P and V2P, while maintaining competitive IS.
Note that we do not report CLIP-Score for PanoDiffusion or DreamCube. PanoDiffusion has no text input, and DreamCube requires six separate per-face text prompts; computing CLIP-Score under our single-prompt setting would be inconsistent and unfair. 
In summary, these results demonstrate that JoPano delivers better panorama generation quality, validating the effectiveness of our unified joint modeling~framework.

\paragraph{Qualitative Results}
We further provide qualitative comparisons to evaluate the effectiveness of our method.
We evaluate both indoor and outdoor scenes and include a stylized example to demonstrate style controllability.
As shown in \cref{fig:T2P compare} and \cref{fig:V2P compare}, JoPano produces panoramas with sharp details, few distortions, improved seam consistency, and good text–image alignment in both T2P and V2P settings.
In the T2P comparison, only PAR and JoPano produce stylized results, and JoPano shows clearer brush strokes and better matches the target style.
In the V2P comparison, Diffusion360 and DreamCube can also generate stylized panoramas, but JoPano better matches both the input view and the requested style.
These results indicate that JoPano preserves the base model’s stylized image generation ability in the panorama setting.

\subsection{Ablation Study}

\paragraph{Position Embedding}
We conduct an ablation study on different types of position embeddings.
Under identical training settings, We compare two types of positional embeddings under identical training settings: (1) UV coordinates on each cubemap face, and (2) our 3D spherical embedding, both encoded with RoPE. in terms of quantitative image quality.
As shown in \cref{tab:rope_comparison}, the spherical positional embedding outperforms the UV positional embedding, as the spherical representation better captures the geometry of~panoramas.

\begin{table}[t]
\centering
\caption{Image quality comparison of different positional encoding (PE) types on SUN360 dataset.}
\begin{tabularx}{\linewidth}{l l *{3}{>{\centering\arraybackslash}X}}
\toprule
Task & PE Type & FID $\downarrow$ & \text{CF} $\downarrow$  & IS $\uparrow$ \\
\midrule
\multirow{2}{*}{T2P} 
                & UV      & 52.26         & 14.15          & 9.86      \\
                & Sphere  & \textbf{48.72} & \textbf{13.13} & \textbf{10.21}      \\
\midrule
\multirow{2}{*}{V2P}
                & UV      & 21.32         & 5.95         & 8.96    \\
                & Sphere  & \textbf{19.97} & \textbf{5.87} & \textbf{9.09}        \\
\bottomrule
\end{tabularx}

\label{tab:rope_comparison}
\end{table}

\begin{table}[t]
\centering
\caption{Comparison of seam consistency on SUN360 dataset.}
\begin{tabularx}{\linewidth}{lcc}
\toprule
Method             & Seam-SSIM $\uparrow$ & Seam-Sobel $\downarrow$ \\ \midrule
Noise              & 0.004               & 129.62    \\
Ground Truth       & 0.847               & 11.16     \\
DreamCube          & 0.725               & 35.85     \\ \midrule
JoPano-T2P w/o CFB & 0.762               & 39.69     \\
JoPano-T2P w/ CFB  & \textbf{0.831}      & \textbf{12.66}     \\\midrule
JoPano-V2P w/o CFB & 0.786               & 41.16     \\
JoPano-V2P w/ CFB  & \textbf{0.861}      & \textbf{12.18}     \\ \bottomrule
\end{tabularx}

\label{tab:seam_ssim}
\end{table}

\begin{figure*}[t!]
    \centering
    \includegraphics[width=\textwidth]{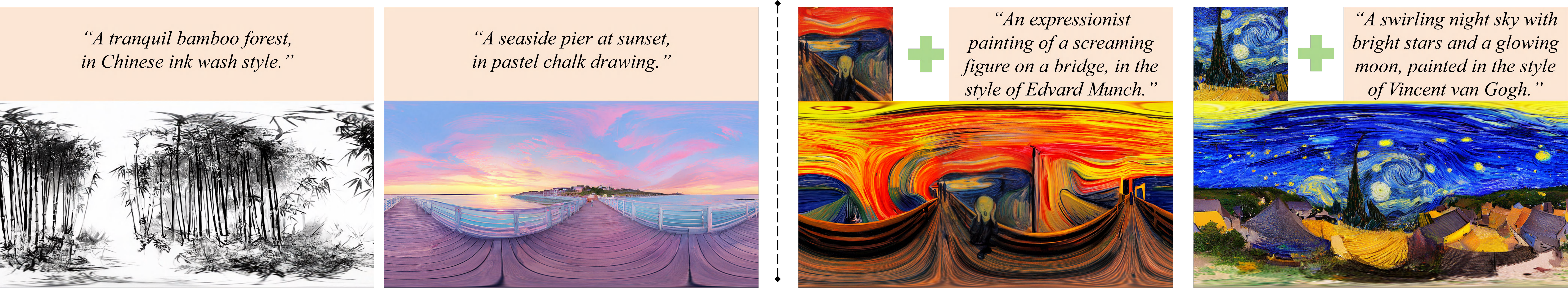}
    \caption{Style panorama generation. T2P generates panoramas from text descriptions that include style, while V2P generates panoramas from stylized view conditions.}
    \label{fig:stylization}
\end{figure*}

\begin{figure*}[h]
    \centering
    \includegraphics[width=\textwidth]{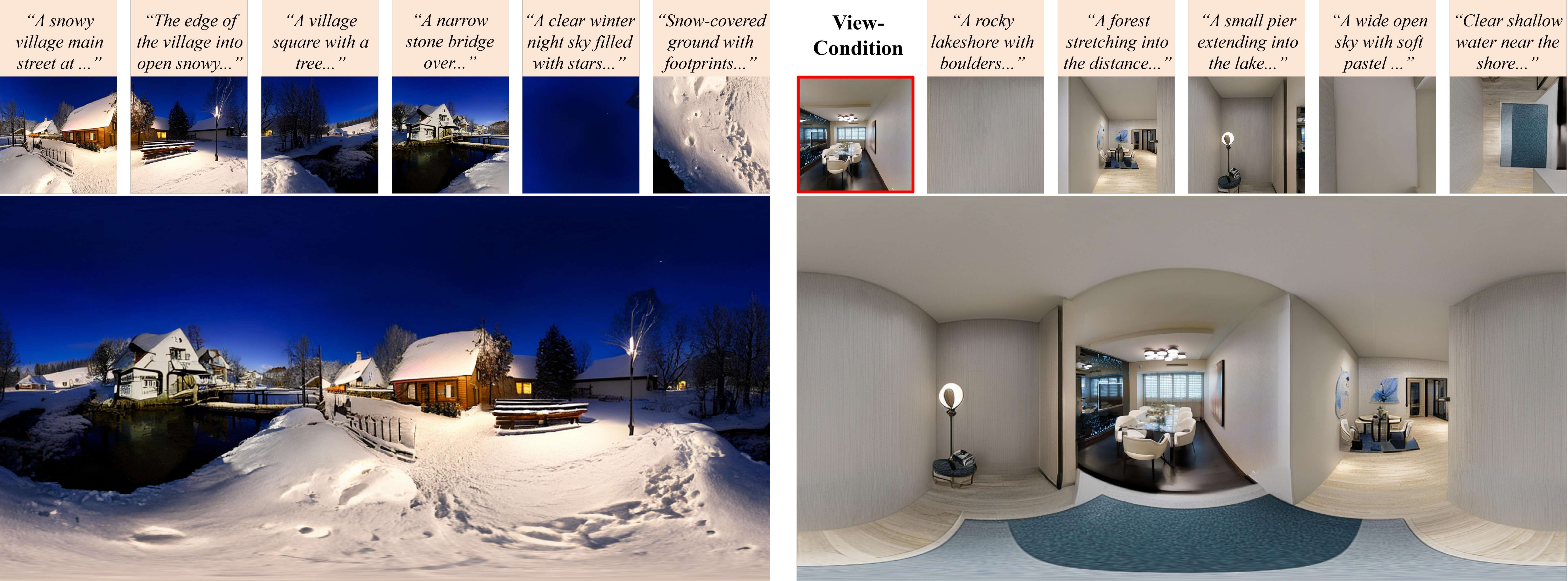}
    \caption{Multi-text generation. In T2P, JoPano generates a panorama from six different text descriptions, while in V2P it generates from one view condition and five text descriptions.
    }
    \label{fig:multi-text prompting}
\end{figure*}

\paragraph{Seam Consistency Comparison}
Cubemap panoramas are prone to seams at face boundaries. We evaluate these artifacts using our proposed metrics and assess the effect of Cross-Face Blending (CFB).
To validate the metrics, we compute them on pure noise cubemaps and on the ground truth cubemaps converted from ERP panoramas in the SUN360 test set. As shown in \cref{tab:seam_ssim}, noise cubemaps obtain very poor scores on both metrics, while ground truth achieves high Seam-SSIM (0.847) and low Seam-Sobel (11.16), confirming that the metrics correlate with seam smoothness.
We then evaluate JoPano. Models with CFB outperform both their versions without CFB and DreamCube on the two metrics. In particular, JoPano-V2P with CFB nearly matches the ground truth seam quality.
These results, together with the visual examples in \cref{fig:CFB}, indicate that CFB effectively improves seam consistency.

\subsection{Additional Experiment}
To further assess the generative generalization capability of JoPano, we conduct two zero-shot experiments: stylized panorama generation and multi-text generation.

\paragraph{Stylized Panorama Generation}
We evaluate stylized panorama generation by using artistic text prompts that do not appear in the training data.
Thanks to the pretrained Sana-DiT backbone, JoPano can follow these prompts with style while still producing panoramas with consistent global structure.
As shown in \cref{fig:stylization}, our method preserves Sana’s stylized image generation ability and extends it to the panorama domain without breaking the underlying scene~layout.

\paragraph{Multi-Text Generation}
We train the model using only a single text prompt per panorama; nevertheless, it generalizes to the multi-text setting.
We show T2P and V2P results with multi-text prompting in \cref{fig:multi-text prompting}. In the T2P setting, each of the six cubemap faces is generated from its own text prompt, and in the V2P setting, one face is given as a view condition and the remaining five faces are generated from five text prompts. In both cases, \MODEL produces a single coherent panorama.

\section{Conclusion}
\label{sec:conclusion}

In this paper, we present JoPano, a joint-face panorama generation framework built on a DiT-based model.
We extend the pretrained Sana backbone with a Joint-Face Adapter that jointly models all six cubemap faces and transfers Sana’s image generation capability to the panorama domain.
In addition, we introduce a condition-switching mechanism that unifies the two key tasks in panorama generation, text-to-panorama and view-to-panorama, within a single diffusion process.
These components enable high-quality panorama generation while handling both tasks in a single, efficient model.
To validate the effectiveness of our approach, we compare JoPano with existing methods on both text-to-panorama and view-to-panorama tasks. 
The experimental results show that our model not only unifies these two tasks within a single framework, but also achieves superior quantitative performance and better visual quality than prior~methods.

\newpage     
{
    \small
    \bibliographystyle{ieeenat_fullname}
    \bibliography{main}
}
\appendix
\clearpage
\setcounter{page}{1}
\maketitlesupplementary

\begin{table*}[h]
\centering
\caption{
Quantitative evaluation at 1024×6 and 512×6 resolutions measured by FID, CLIP-FID (CF), IS, and CLIP-Score (CS).
}
\begingroup
\begin{tabularx}{\textwidth}{l *{4}{>{\centering\arraybackslash}X} *{4}{>{\centering\arraybackslash}X}}
\toprule
\multirow{2}{*}[-0.7ex]{Methods} & \multicolumn{4}{c}{SUN360} & \multicolumn{4}{c}{Structure3D} \\
\cmidrule(lr){2-5}\cmidrule(lr){6-9}
& FID $\downarrow$ & \mbox{CF} $\downarrow$ & IS $\uparrow$ & \mbox{CS} $\uparrow$
& FID $\downarrow$ & \mbox{CF} $\downarrow$ & IS $\uparrow$ & \mbox{CS} $\uparrow$ \\
\midrule
JoPano-T2P (1024×6) & \textbf{28.58} & 12.37 & \textbf{7.97} & 30.04 & \textbf{32.70} & \textbf{15.71} & 3.25 & \textbf{28.52} \\
JoPano-T2P (512×6)  & 29.83 & \textbf{10.95} & 7.80 & \textbf{30.12} & 34.44 & 16.17 & \textbf{3.51} & 27.96 \\
\midrule
JoPano-V2P (1024×6) & \textbf{12.93} & 4.14 & \textbf{7.27} & 27.84 & \textbf{16.28} & \textbf{2.93} & \textbf{3.08} & \textbf{27.55} \\
JoPano-V2P (512×6)  & 13.07 & \textbf{4.06} & 7.05 & \textbf{27.93} & 16.75 & 3.97 & 3.04 & 27.33 \\
\bottomrule
\end{tabularx}

\vspace{5mm}
\endgroup
\label{tab:4k}
\end{table*}

\begin{figure*}[ht]
    \centering
    \includegraphics[width=\linewidth]{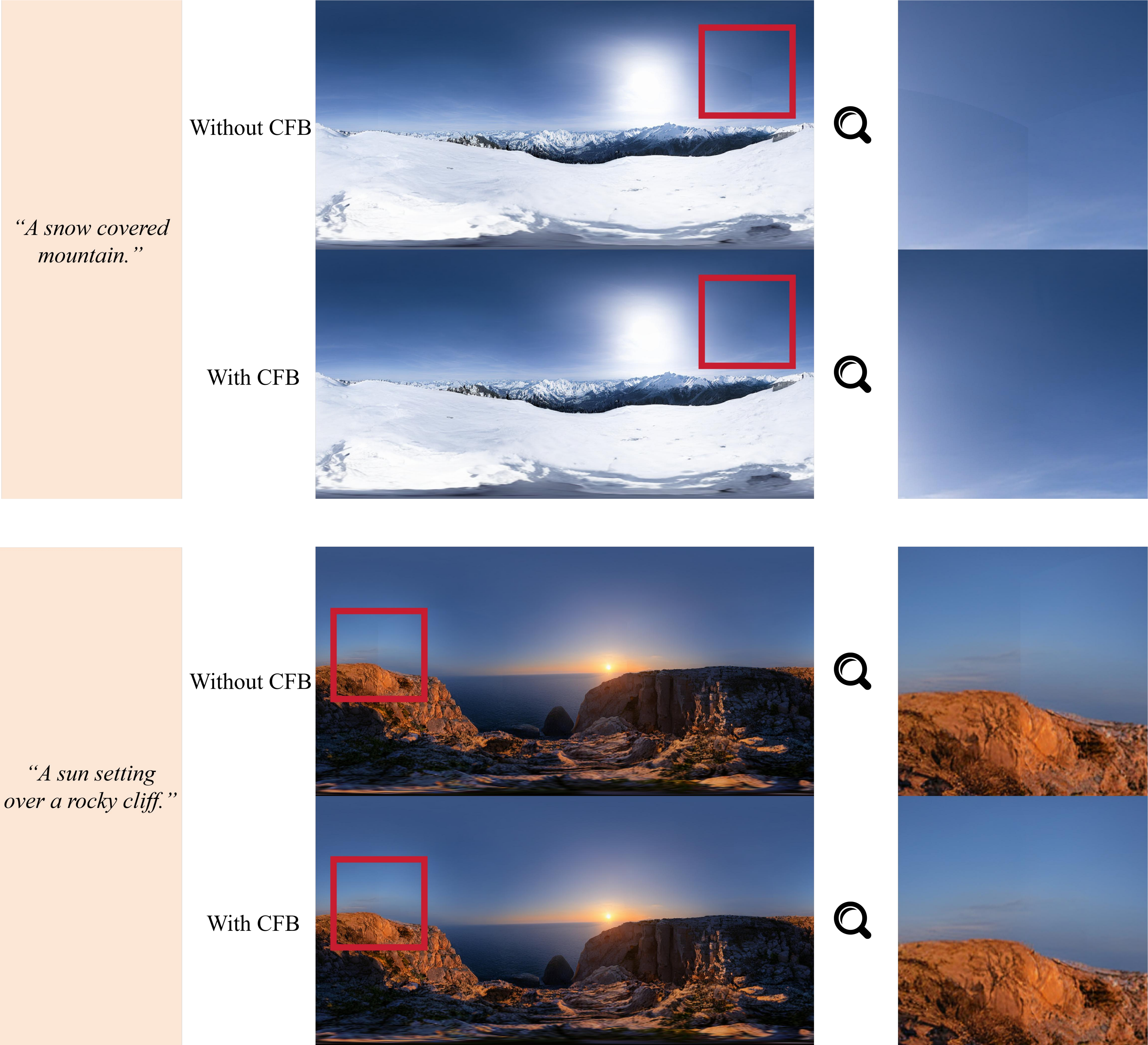}
    \caption{More results of ERP panoramas with and without Cross-Face Blending (CFB).}
    \label{fig:cfb_suppl}
\end{figure*}

\begin{figure*}[ht]
    \centering
    \includegraphics[width=\linewidth]{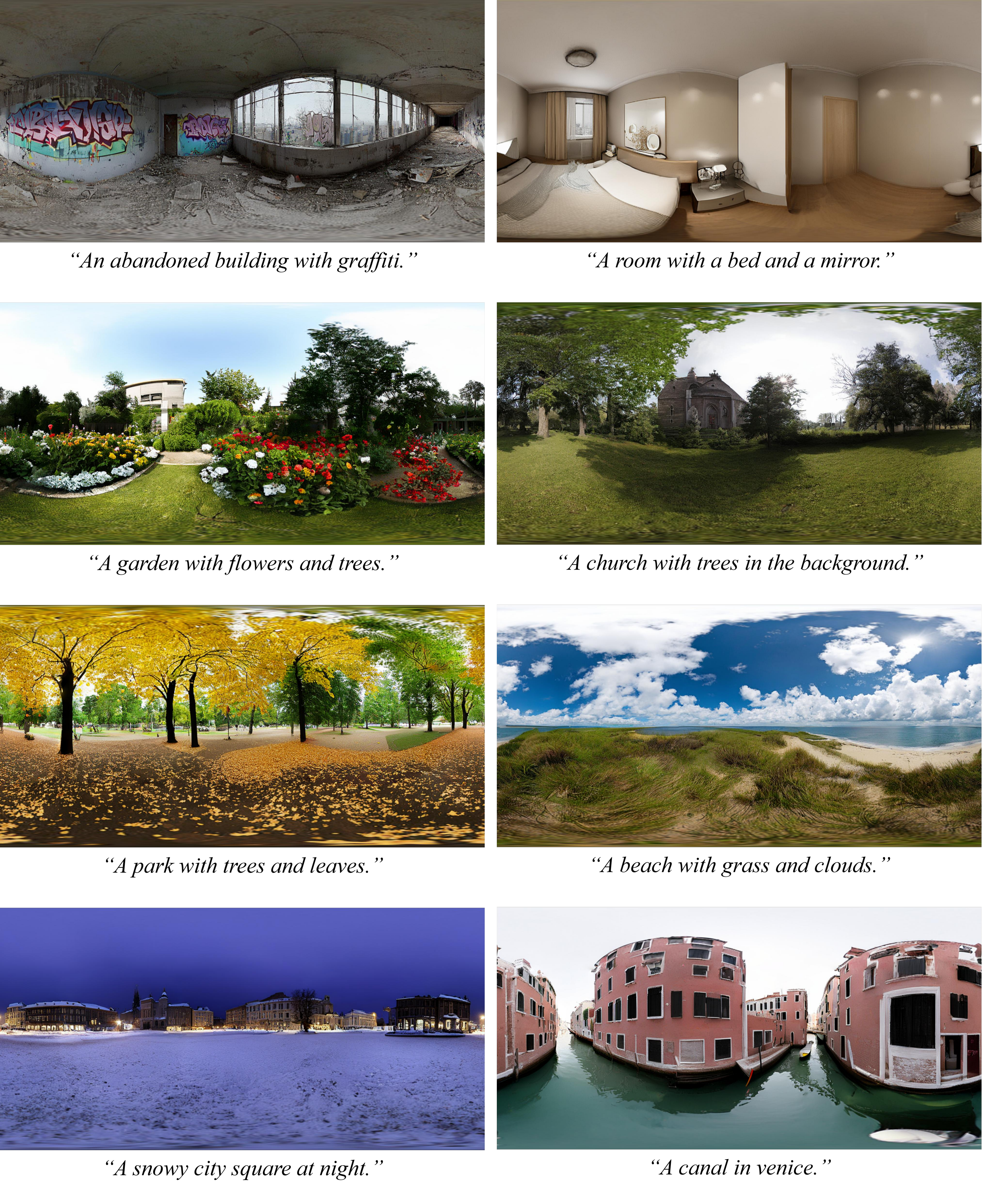}
    \caption{More results of T2P generation.}
    \label{fig:suppl_t2p}
\end{figure*}

\begin{figure*}[ht]
    \centering
    \includegraphics[width=\linewidth]{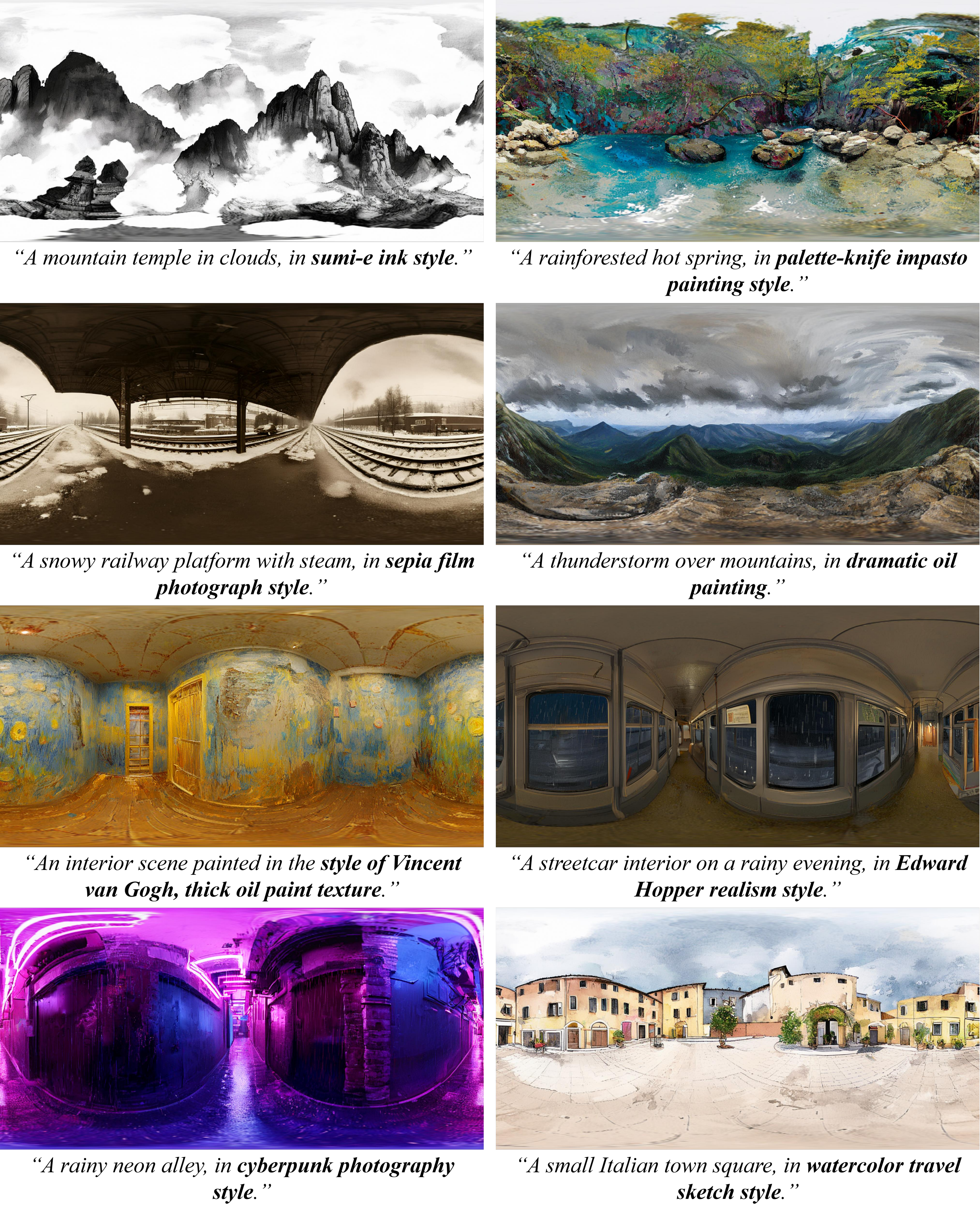}
    \caption{More results of stylized T2P generation.}
    \label{fig:suppl_t2p_style}
\end{figure*}

\begin{figure*}[ht]
    \centering
    \includegraphics[width=\linewidth]{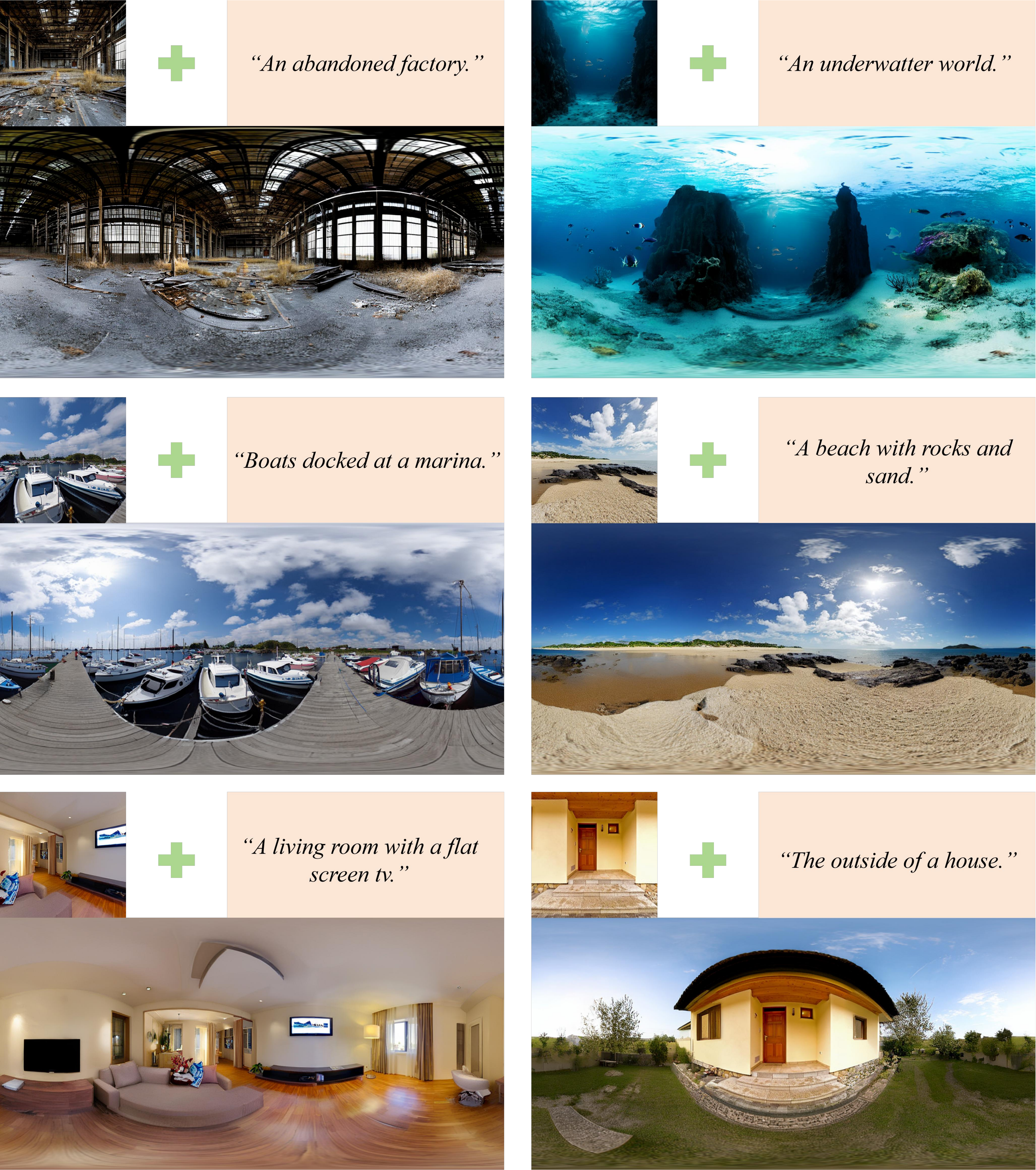}
    \caption{More results of V2P generation.}
    
    \vspace{10mm}
    \label{fig:suppl_v2p}
\end{figure*}

\begin{figure*}[ht]
    \centering
    \includegraphics[width=\linewidth]{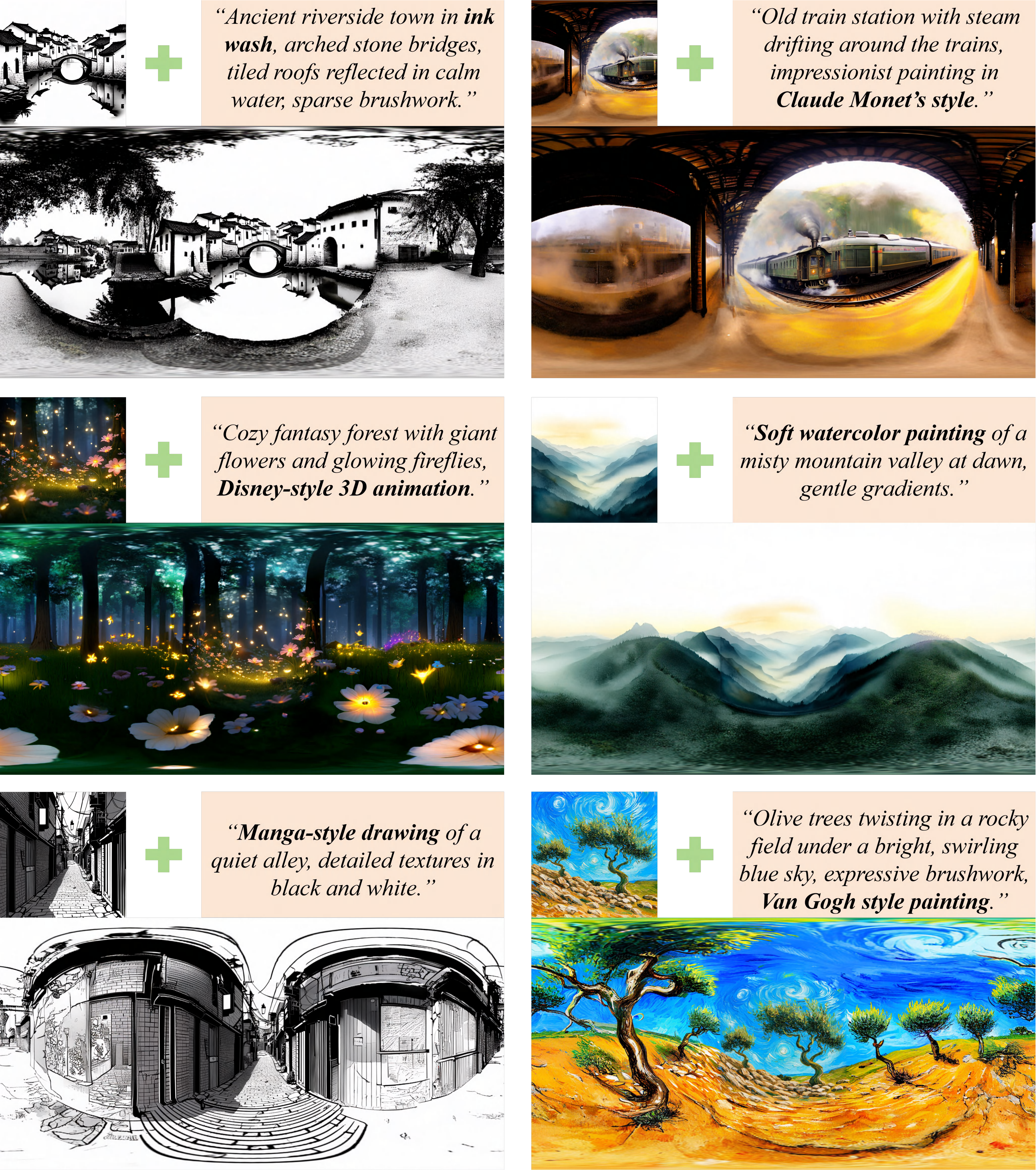}
    \caption{More results of stylized V2P generation.}
    \vspace{10mm}
    \label{fig:suppl_v2p_style}
\end{figure*}

\section{Details of Cross-Face Blending}
\subsection{Implementation Details}

In this section, we provide implementation details of Cross-Face Blending (CFB), whose Poisson formulation is given in \cref{eq:poisson}.
For each cubemap face $g_i$, we discretize the domain $\Omega_i$ on the image grid and solve for a discrete unknown
$f_i(u,v)$ defined on pixels $(u,v)$.
We first compute the discrete Laplacian of $g_i$ using the standard 5-point stencil:
\begin{align}
\Delta g_i(u,v) &=g_i(u+1,v) + g_i(u-1,v) + g_i(u,v+1)\notag\\
&\quad+ g_i(u,v-1) - 4g_i(u,v).
\end{align}
For interior pixels, the Poisson equation is discretized as
\begin{align}
f_i&(u+1,v) + f_i(u-1,v)+ f_i(u,v+1) \notag\\
&+ f_i(u,v-1)-4 f_i(u,v)= \Delta g_i(u,v).
\end{align}
The Dirichlet boundary values on $\partial\Omega_i$ are fixed to the pixelwise averages of $g_i$ and its neighboring faces along each shared edge.
To solve this linear system, we apply the iterative Gauss–Seidel method.
Starting from the initial guess $f_i^{(0)} = g_i$, we update interior pixels sequentially.
At iteration $k+1$, the value at pixel $(u,v)$ is updated using the most recent values of its neighbors:
\begin{align}
f_i^{(k+1)}(u,v)
&= \frac{1}{4}\Big(
    f_i^{(k)}(u+1,v)
  + f_i^{(k+1)}(u-1,v) \notag\\
&\quad
  + f_i^{(k)}(u,v+1)
  + f_i^{(k+1)}(u,v-1) \notag\\
&\quad
  - \Delta g_i(u,v)
\Big),
\end{align}
where the left and top neighbors have already been updated at iteration $k+1$, while the right and bottom neighbors use values from iteration $k$.
In practice, we run 200 iterations in all our experiments to obtain the blended solution $f_i$ for each face.

\subsection{Visual Effect of Cross-Face Blending}
We further illustrate the visual effect of Cross-Face Blending on ERP panoramas. \cref{fig:cfb_suppl} compares results with and without CFB, showing that it produces smoother cross-face transitions and reduces seam artifacts.

\section{Experimental Details of Compared Methods}
In this section, we describe how we reproduce DreamCube~\cite{huang2025dreamcube} for comparison.
We follow the data processing pipeline described in the original paper on our datasets.
For datasets that are not provided in cubemap format, we first convert ERP panoramas into cubemaps using standard perspective projection.
We then use BLIP-2~\cite{li2023blip} to generate an image caption for each cube face.
Next, to annotate the depth of these panoramas, we build a high resolution panorama depth estimation pipeline by connecting the panorama depth estimation work Depth Anywhere~\cite{wang2024depth} and the image-guided depth upsampling work PromptDA~\cite{chen2023promptda}, which supports panorama depth estimation.
We use this pipeline to perform depth estimation on ERP panoramas and then project the obtained depth panoramas into cubemaps.
After these preprocessing steps, we run DreamCube with a single view image, its corresponding depth map, and six separate text prompts.

\vfill
\section{Additional High Resolution Experiment}
We further validate the high resolution generation capability of JoPano by retraining the model at a resolution of $1024 \times 1024 \times 6$, and visualizing the outputs in ERP format at $4096 \times 2048$. From the comparisons in ~\cref{tab:4k}, we evaluate FID, CLIP-FID, IS, and CLIP-Score. The results prove that our method maintains strong performance even in this high resolution setting.

\vfill
\section{More Panorama Results}
We present additional results in \cref{fig:suppl_t2p,fig:suppl_t2p_style,fig:suppl_v2p,fig:suppl_v2p_style} to further illustrate the performance of JoPano on panorama generation.
These examples show that JoPano consistently produces sharp, low-distortion panoramas with improved seam consistency, faithful text alignment, and high-quality stylized results in both T2P and V2P settings.

\vfill
\section{Limitation and Future Work}
Although JoPano achieves promising results, it still exhibits several limitations, mainly for the following two reasons.
First, since the original training images are 1024$\times$512, which are simply resized to 2048$\times$1024 or 4096$\times$2048 during training, the generated panoramas exhibit noticeable blurriness in fine details in fine details.
Second, although Sana is an efficient DiT model with reduced memory consumption and fast inference, it still lags behind Flux in terms of visual quality.
In future work, we plan to enhance panorama generation quality from both the data and model perspectives: on the one hand, by constructing a large-scale, high-resolution dataset, and on the other hand, by using Flux as the base model to train the Joint-Face Adapter, thereby achieving higher-quality panorama generation.

\end{document}